\documentclass{article}
\usepackage{graphicx}
\usepackage{xcolor}%
\usepackage{amsmath} 
\usepackage{appendix}

\usepackage{booktabs}
\usepackage{hyperref}
\usepackage{multirow}
\begin{document}
\addtolength{\textwidth}{2.75in}
\addtolength{\topmargin}{-0.875in}
\addtolength{\textheight}{1.75in}

\title{\textbf{Deep Neural Networks as Discrete Dynamical Systems: Implications for Physics-Informed Learning}}

\author{
  Abhisek Ganguly\thanks{Corresponding author: \href{mailto:a@jncasr.ac.in}{abhisek@jncasr.ac.in}}\\
  Engineering Mechanics Unit,\\
  Jawaharlal Nehru Centre for Advanced Scientific Research,\\
  Jakkur, Bangalore, 560064, Karnataka, India
  \and
  Santosh Ansumali\\
  Engineering Mechanics Unit,\\
  Jawaharlal Nehru Centre for Advanced Scientific Research,\\
  Jakkur, Bangalore, 560064, Karnataka, India
  \and
  Sauro Succi\\
  Italian Institute of Technology,\\
  Viale Regina Elena 291, 00161 Rome, Italy\\
  University of Roma Tre, Italy\\
  Physics Department, Harvard University, Cambridge, USA\\
  Cornell University, Ithaca, USA
}

\maketitle

%---------------------------------------------------------------------------------%
\section{Abstract}
%---------------------------------------------------------------------------------%
We revisit the analogy between feed-forward deep neural networks (DNNs) and discrete dynamical systems derived from neural integral equations and their corresponding partial differential equation (PDE) forms.
A comparative analysis between the numerical/exact solutions of the Burgers' and Eikonal
equations, and the same obtained via PINNs is presented. We show that PINN learning provides a different computational pathway compared to standard numerical discretization in approximating essentially the same underlying dynamics of the system.
Within this framework, DNNs can be interpreted as discrete dynamical systems whose layer-wise evolution approaches attractors, and multiple parameter configurations may yield comparable solutions, reflecting the degeneracy of the inverse mapping. In contrast to the structured operators associated with finite-difference (FD) procedures, PINNs learn dense parameter representations that are not directly associated with classical discretization stencils. This distributed representation generally involves a larger number of parameters, leading to reduced interpretability and increased computational cost. However, the additional flexibility of such representations may offer advantages in high-dimensional settings where classical grid-based methods become impractical.

%---------------------------------------------------------------------------------%
\section{Introduction}
%---------------------------------------------------------------------------------%
Deep neural networks (DNNs) have emerged as powerful tools for approximating complex functional relationships across a wide range of scientific applications, including the solution of partial differential equations (PDEs) through Physics-Informed Neural Networks (PINNs)~\cite{Raissi2019, karniadakis2021physics}. Traditional PINNs incorporate governing equations directly into the Loss function, enabling the solution of forward and inverse PDE problems without a reliance on labeled data. Recent studies in the chemical physics literature have also explored PINNs for solving physically relevant systems, including applications to stiff chemical kinetics and molecular electrostatics~\cite{ji2021stiff, achondo2025pinn}. Alongside their practical success, there is a growing interest in developing theoretical frameworks that can provide a deeper understanding of their learning dynamics and representational properties~\cite{Kovachki2023}.

In this work, we revisit the analogy between feed-forward DNNs and discrete dynamical systems in relaxation form, originating from the discretization of neural integral equations~\cite{zappala2022neural}. Within this framework, the propagation of features across layers can be interpreted as a time-stepping process, in which the network evolves toward attractors that may, in principle, correspond to learned representations. This viewpoint provides a natural connection between deep learning architectures and classical concepts from numerical analysis and computational physics.

This perspective differs from, but is related to, the framework of neural ordinary differential equations~\cite{chen2018neural,he2016deep}, where the depth of the network is interpreted as a continuous time variable. Here, instead, we emphasize a formulation based on integral representations and their discrete counterparts, which leads to relaxation-type dynamics and potentially nonlocal interactions in feature space.

The main objective of this work is to investigate the learning dynamics of deep neural networks, and in particular PINNs, through the lens of discrete dynamical systems. This perspective highlights the existence of multiple parameter configurations leading to essentially the same learned solution, revealing a fundamental non-uniqueness in the mapping between network parameters and physical outputs. For context, we contrast this learning behavior with classical numerical discretization methods, such as finite-difference (FD) schemes~\cite{leveque2007finite}, which typically rely on structured and more directly interpretable operators.

Understanding this connection is of interest for several reasons. First, it provides insight into the relationship between network architecture and solution accuracy when learning PDEs~\cite{E2017dynamical}. Second, it offers a route toward improving the interpretability of machine learning models in scientific computing. While universal approximation results guarantee the expressive power of neural networks~\cite{cybenko1989approximation, hornik1989multilayer, hornik1991approximation}, they do not address how such representations are constructed across layers. From the present viewpoint, this process can be interpreted as the evolution of a dynamical system towards a target state through a sequence of transformations.

Bringing this connection to light contributes to the broader goal of establishing a theoretical foundation for machine learning methods in computational physics, and may help guide the development of more reliable and efficient architectures for scientific applications.

%---------------------------------------------------------------------------------%  
\section{Physics-informed neural networks (PINNs):}
%---------------------------------------------------------------------------------%
The basic idea of deep neural networks (DNNs)~\cite{Lecun2015} is to represent a given $d$-dimensional 
output $y$ through the recursive application 
of a nonlinear and nonlocal map to the input data $x$.

\begin{figure}
\centering
\includegraphics[scale=0.5]{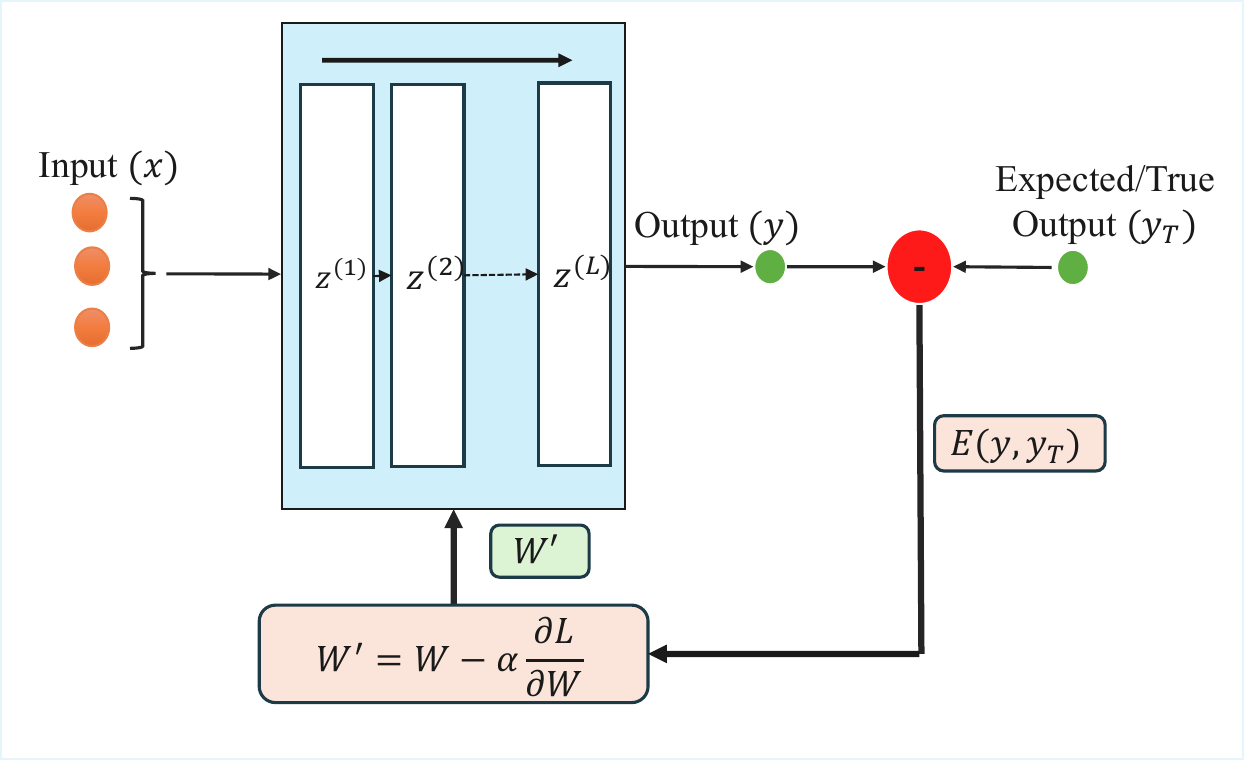}
\caption{The basic blocks of the DNN architecture.
The procedure is repeated over a set of collocation points in the domain. For PINNs, Loss function $E$ is given by Eq.~\ref{Loss}, where $y_T = u_{\text{data}}$ for an optional, data-driven architecture.
} 
\label{fig:DNN}
\end{figure}

For a DNN consisting of an input layer $x$,  $L$ hidden layers
$z^{(1)} \dots z^{(L)}$, each containing $N$ neurons, and an output layer $y$,
the update chain $x \to z^{(1)} \dots \to z^{(L)} \to  y$
reads symbolically as follows:
\begin{equation}\label{eq:gen_dnn}
\begin{aligned}
z^{(0)} &= x,\\
z^{(1)} &= f(W^{(1)} z^{(0)} - b^{(1)}),\\
&\;\vdots\\
z^{(L)} &= f(W^{(L)} z^{(L-1)} - b^{(L)}),\\
z^{(L+1)} &= f(W^{(L+1)} z^{(L)} - b^{(L+1)}) = y.
\end{aligned}
\end{equation}
where $W^{(l)}$ are weight matrices, $b^{(l)}$ are bias vectors, and $f$ is a nonlinear activation function. A schematic is shown in Fig.~\ref{fig:DNN}.

PINNs are a class of deep neural networks designed to approximate solutions of partial differential equations (PDEs), which arise across a wide range of fields including physics, engineering, and chemical kinetics.  In 1D PINNs, the input and output are taken in the form of 
scalar functions of a variable $q$, representing a coordinate in feature space
(e.g. space or space-time). From here onwards, we represent physical space and time using the notations $\mathcal{X}$ and $\mathcal{T}$ respectively. 

The network output $y(q)$ is interpreted as an approximation 
to a physical field $u(q)$. This is a simplified view of the DNN described in Eq.~\ref{eq:gen_dnn}. Furthermore, general DNN architectures may incorporate layerwise normalization, a feature that is not typically employed in standard PINN formulations. We note here that the dimensions of $z$ for any layer need not be the same as the input/output dimensions. We look at the PINNs that have same number of neurons per layer in the hidden layers.

In contrast to standard supervised learning, the target function is not assumed to be known 
over the entire domain. Instead, the network is trained to satisfy a set of physical constraints, 
typically expressed in terms of a differential operator $\mathcal{N}$. These include initial 
and/or boundary conditions, along with the governing equations of the system.

A data-driven training procedure minimizes a composite Loss function of the form
\begin{equation}\label{Loss}
E[W] = E_{\text{data}} + E_{\text{physics}},
\end{equation}
where
\begin{equation}
E_{\text{data}} = ||u(q) - u_{\text{data}}(q)||,
\end{equation}
enforces agreement with available data, and
\begin{equation}
E_{\text{physics}} = \|\mathcal{N}\left[u(q)\right]\| = \|\lambda_1\mathcal{N}_{PDE}\left[u(q)\right]+\lambda_2\mathcal{N}_{IC}\left[u(q)\right]+\lambda_3\mathcal{N}_{BC}\left[u(q)\right]\| ,
\end{equation}
penalizes the residual of the governing equations (PDEs, initial and boundary conditions). Note that $E_{\text{data}}$ is not included for systems that are not data-driven. Also, $\lambda_1, \lambda_2, \lambda_3$ are generally some constant weights, signifying the importance of each constraint in the optimization.

The required spatial and temporal derivatives entering $\mathcal{N}$ are computed 
via automatic differentiation through the network.

The minimization of the Loss function is typically performed 
via a steepest descent procedure (or stochastic variants thereof), 
\begin{equation}
W' = W - \alpha \frac{\partial E}{\partial W},
\end{equation}
where $\alpha$ is a relaxation parameter controlling the convergence 
(learning rate) of the procedure.

Convergence of the learning procedure is associated with the 
attainment of local minima of the Loss function. 
Given the large dimensionality of the weight space $\mathcal{W}$ and the 
complex structure of the Loss landscape, the number
of local minima is generally very large and it is difficult to 
identify a priori those corresponding to physically meaningful solutions.

% --------------------------
% --------------------------

The basic idea of the entire procedure is to construct a general 
input-output relation 
\begin{equation}
\label{FUN}
y(q)=F_f[x(q);W,b],
\end{equation}
such that the resulting function satisfies the prescribed physical constraints.
In the PINN framework, the functional mapping $F_f$ is trained such that
\begin{equation}
\mathcal{N}[y(q)] \approx 0,
\end{equation}
thereby embedding the governing equations directly into the learning process.

It is to be noted that the functional relation in Eq.~\ref{FUN} depends 
on a large set of parameters, namely the weights and biases: for a chain of $L$ 
hidden layers, each made of $N$ fully connected neurons, the number of free parameters 
scales as $P \sim \mathcal{O}\left(L N^2\right)$ for large $N$ and $L$. This large parametric freedom enables the network 
to approximate complex functional dependencies, while the inclusion of physical constraints 
serves to regularize the solution space and guide the learning towards physically admissible solutions.

In this work, we build upon this framework to expose the analogy between PINNs (from the viewpoint of feed-forward 
deep neural networks) and discrete dynamical systems based on neural integral equations.

%---------------------------------------------------------------------------------%
\section{DNNs as discrete dynamical systems}\label{sec:DDS}
%---------------------------------------------------------------------------------%
\subsection{Time-like analogy of DNN layers}
In recent years,  Neural Ordinary Differential Equations (Neural ODEs~\cite{chen2018neural}) have emerged as  
a powerful paradigm in deep learning, one that combines continuous-time dynamical 
systems with neural networks. 
The field started with the so called ResNet model~\cite{he2016deep} where it was proposed 
that a better layer-to-layer update is given by: 
\begin{equation}
    z^{(l+1)}=z^{(l)}+  F(z^{(l)},\epsilon).
\end{equation}
where $\epsilon$ is a suitable smallness parameter.
It is quickly recognized that this is just a forward Euler discretization of the  
underlying ODE, $\dot z = f(z)$, with $F(z,\epsilon)= \epsilon f(z)$, $\epsilon$
being the timestep.  

Thus, the model was refined as a neural ODE model which views  
the evolution of hidden states in a neural network as the solution to 
an ordinary differential equation (ODE), where the derivative of the hidden 
state with respect to time (or depth) is parametrized by a neural network:
\begin{equation}\label{eq:node}
    \frac{\mathrm{d}z(t)}{\mathrm{d}t} = f_{\theta}\big(z(t), t\big),
\end{equation}
with $f_{\theta}$ being the neural network with parameters $\theta$. 
Instead of stacking discrete layers as in traditional ResNets, the output at any ``depth'' $t\, (\equiv l)$ 
is obtained by solving the ODE from an initial state $z(0)$ using numerical solvers (e.g., Euler). This continuous-depth approach offers several advantages: memory efficiency (via the adjoint sensitivity method for backpropagation, achieving $\mathcal{O}(1)$ memory cost~\cite{chen2018neural}), adaptive computation 
(solvers can adjust step size based on error tolerance), the ability to handle irregular 
time-series data, and most important, a direct connection to dynamical systems theory.

\subsection{Discrete dynamical systems in relaxation form}
We interpret the feature vectors $z$ as signals depending on the input features $x=x(q)$, whose evolution across layers can be described in terms of neural differential equations. Within this perspective, standard concepts of stability and convergence from numerical analysis naturally apply to deep neural networks, including the PINN framework~\cite{tucny2026randomness}. In particular, several neural architectures explicitly leverage such principles to ensure stable training dynamics, notably through implicit or equilibrium formulations~\cite{DEQ,haber2017stable,chen2018neural}.

As discussed above, the layer index of a deep neural network may be interpreted as a discrete time variable. In this spirit, the feature vector $z$ can be promoted to a continuous field $z(q,t)$ depending on both a spatial (or feature) coordinate $q$ and a depth-representative time $t$~\cite{SucciNPDE2025}.
Within this framework, the standard DNN update can be interpreted as a first-order relaxation dynamics toward a nonlinear attractor as follows:
\begin{equation}
\label{REL}
\partial_t z = -\gamma (z - z^{att}),
\end{equation}
where $z(q,t)$ is the signal and $z^{att}(q,t)$ is a local attractor.
The parameter $\gamma$ controls the relaxation rate. Note that this formulation is a dynamical system formulation of a more general Eq.~\ref{eq:node}.

The attractor is defined through a nonlinear transformation of a linearly
convoluted signal:
\begin{equation}
z^{att}(q,t) = f\left(Z(q,t)\right),
\end{equation}
with
\begin{equation}\label{eq:activation_integral}
Z(q,t) = - b(q,t) + \int W(q,q',t) z(q',t)\, dq',
\end{equation}
where $W(q,q',t)$ is a convolution kernel at time $t$ and $b(q,t)$ is a bias term.
The function $f$ represents a local nonlinear activation applied on $Z$.

%--------------------------------------------------------------------------%
\begin{figure}[htbp]
\centering
\includegraphics[width=0.95\linewidth]{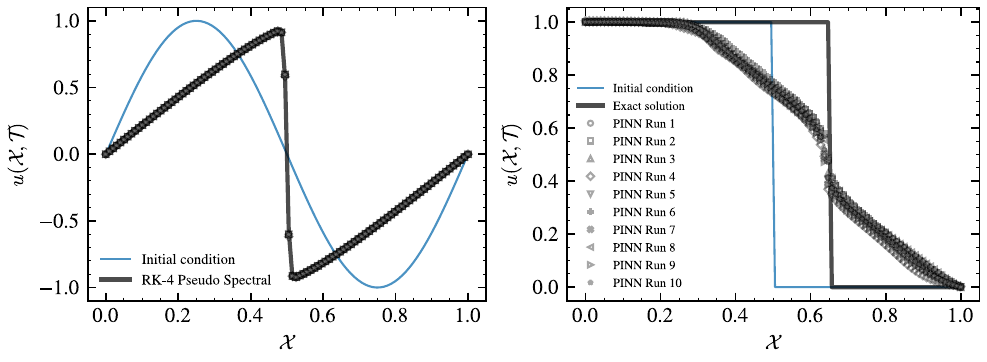}
\caption{Comparison of PINN solutions for the 1D Burgers' equation. 
\textbf{Left:} Viscous case ($\nu > 0$) at $t=0.3\,\mathrm{s}$, where solutions from 10 independent PINN runs are plotted against the FDM pseudo-spectral solution, showing good agreement. 
\textbf{Right:} Inviscid case ($\nu = 0$), where the same PINN architecture fails to learn the correct solution in both the independent runs.}
\label{fig:pinn_vs_dyn}
\end{figure}
%--------------------------------------------------------------------------%

The dynamics in Eq.~\ref{REL} describes a relaxation of the signal $z$
towards the attractor $z^{att}$, which itself evolves through the
combined action of nonlocal linear mixing (via $W$) and local nonlinear
deformation (via $f$). The interplay of these two operations underlies
the expressive power of deep neural networks.

The steady states of the dynamics are given by the fixed-point condition,
\begin{equation}
\label{FP}
z^* = f\left(W z^* - b\right),
\end{equation}
which may admit multiple solutions depending on the nonlinearity of $f$
and the choice of $\{W,b\}$. This strategy of finding the equilibrium/attractor for an ``infinite'' depth system is used by Deep Equilibrium Models (DEQ)~\cite{DEQ}, which arises from an ideal scenario that $z(q,t)$ should stop evolving after reaching the true solution of the learning problem.

This representation also highlights the inverse nature of the learning problem, which can be made explicit from the fixed-point relation
\begin{equation}\label{fixed_point}
W z^* - b = f^{-1}(z^*),
\end{equation}
defining a system for the unknown parameters $\{W,b\}$. Since this system is 
generally underconstrained, multiple parameter sets can yield the same fixed point $z^*$. 

A discrete-time version of Eq.~\ref{REL}, obtained via forward Euler
time integration and spatial discretization over $N$ nodes/neurons, reads
\begin{equation}
\label{RELAX}
z_i^{(t+1)} = (1-\omega) z_i^{(t)} + \omega z_i^{att,(t)},
\end{equation}
where $\omega = \gamma \Delta t$ is a relaxation parameter, and $i \in [0,N-1]$.
For $\omega = 1$, this reduces exactly to the standard layerwise
update of a feed-forward neural network (PINN here), with the layer index
playing the role of discrete time.

Accordingly, the DNN propagation can be written in compact form as
\begin{equation}
\label{eq:chain_update}
z^{(t+1)} = T_f[W^{(l+1)},b^{(l+1)};\omega] z^{(t)},
\end{equation}
with propagator
\begin{equation}
\label{eq:propagator}
T_f[W,b;\omega] = (1-\omega) I + \omega f[W,b].
\end{equation}
For $\omega = 1$, Eq.~\ref{eq:chain_update} becomes $z^{(t+1)} = f\left[W^{(t+1)},b^{(t+1)}\right] = f\left(W^{(t+1)}z^{(t)}-b^{(t+1)}\right)$, which is the standard discrete feed-forward DNN update.

From this formulation, it is apparent that the standard implementations provide no guarantee that a finite-layer DNN will relax to \(z^{\text{att}}\) in the usual dynamical sense (owing to an ever evolving attractor), even though it may be beneficial to encourage this relaxation \emph{in addition} to Loss function minimization. In an ideal scenario from an infinite-depth perspective, the steady state \(z = z^{\text{att}} = y_{\text{data}}\) would mark the completion of learning in the discrete dynamical sense, because reaching equilibrium for the underlying neural PDE would then be equivalent to solving the target physical PDE (in the PINN case).

However, the Loss function (Eq.~\ref{Loss}) contains no explicit information about the underlying neural PDE, which therefore evolves in an essentially unconstrained manner as the weights are updated at each optimization step. Consequently, convergence to a fixed point and minimization of the Loss often do not coincide~\cite{SucciNPDE2025}. More precisely, the standard formulation only enforces that trajectories generated by \emph{different} neural PDEs (arising from non-unique trained weights and possibly heading toward different attractors) all pass through the common point \(z(q,t) = y_{\text{data}}\) at \(t = L+1\) for any $q$ via the physics-informed Loss.

Furthermore, like the fixed-point relation (Eq.~\ref{fixed_point}), Loss minimization is highly underdetermined in weight space. This manifests as degenerate representations even when the final network output appears identical. In the subsequent sections we demonstrate how this parameter-space degeneracy appears in practice, producing similar Loss minima that may or may not correspond to the true physical solution, depending on the effectiveness of the optimizer for the specific problem.

%--------------------------------------------------------------------------%
\begin{figure}[t]
    \centering
    \includegraphics[width=\linewidth]{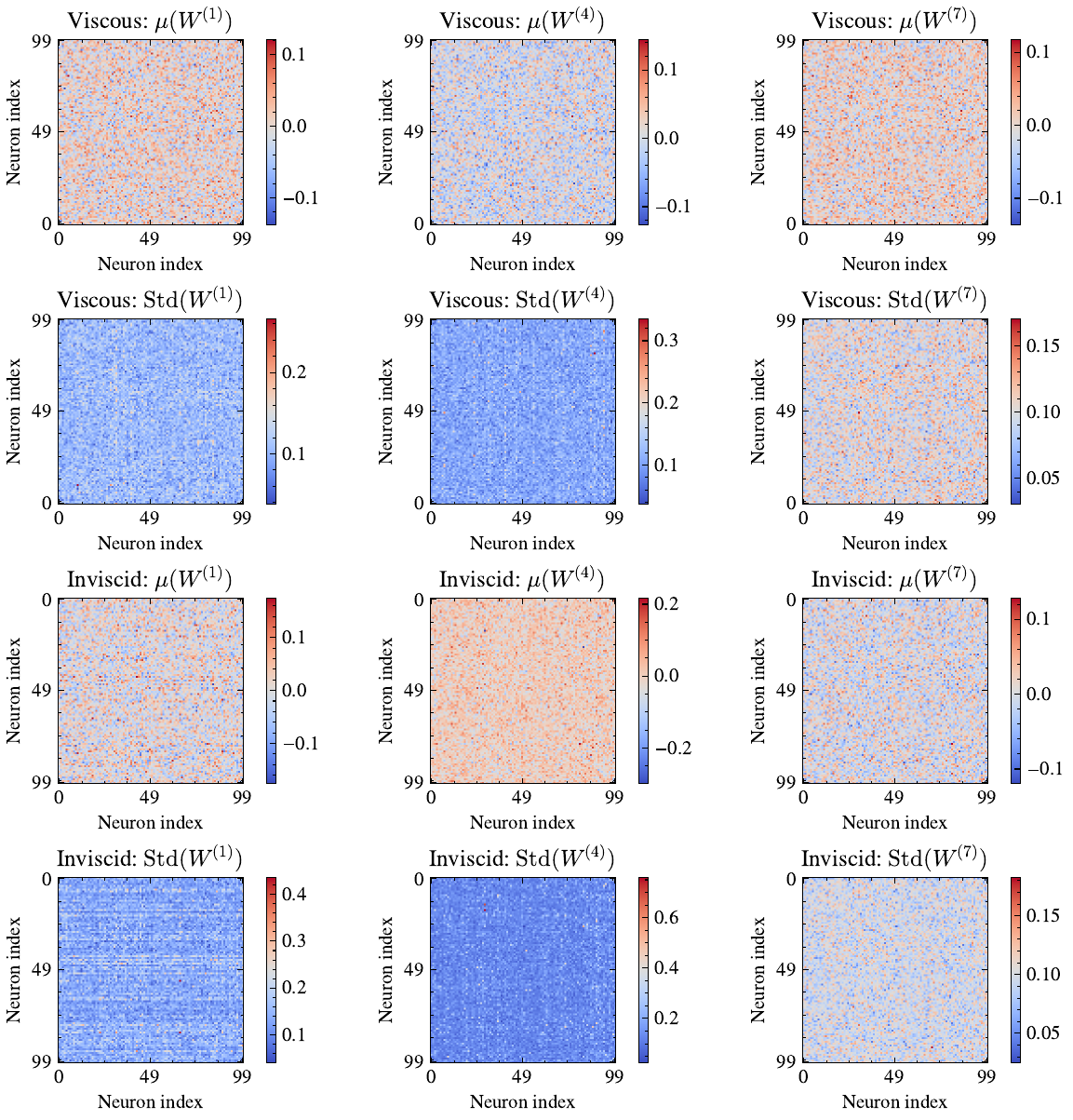}
    \caption{The ensemble weight matrix statistics for layers 1,4 and 7 of the PINN
    solution of the viscous and inviscid Burgers' equation in fluid dynamics for 10 separate, independent runs The mean $\mu(W)$ and standard deviation $Std(W)$ matrices are shown as well, lacking any lower order correlations of significance.}
    \label{fig:weights_burgers_sine} 
\end{figure}
%--------------------------------------------------------------------------%

%---------------------------------------------------------------------------------%
\section{Demonstrating degeneracy in PINNs:}\label{sec:PINN}
%---------------------------------------------------------------------------------%
In this section, we demonstrate the non-uniqueness in the learning paths of PINNs applied to representative PDEs in hydrodynamics. Furthermore, we analyze these results to highlight cases where the obtained solutions align or misalign with the expected physical behavior. We consider the Burgers' (viscous and inviscid) equations and the Eikonal equation, and perform ten independent training procedures for the PINN architecture chosen for each problem. An independent run here corresponds to a training process performed starting from a different set of initial network weights, drawn at random from a uniform distribution using a different seed. All the other training hyperparameters remain the same across each run. The networks are trained over one full convection time for the corresponding problems. Furthermore, the batch size is kept equal to the training data size (i.e., one batch only) in order to keep the non-determinism due to SGD (Stochastic Gradient Descent) to a minimal.
At this stage, we define the norms over $n$ test points as follows, 
\begin{gather}
L_1 = \|\mathbf{e}\|_{1} = \frac{1}{n}\sum_{i=1}^{n} |e_i|, \nonumber \\
L_2 = \|\mathbf{e}\|_{2} = \left( \frac{1}{n}\sum_{i=1}^{n} e_i^{2} \right)^{1/2}, \nonumber \\
L_\infty = \|\mathbf{e}\|_{\infty} = \max_{1 \le i \le n} |e_i|, \nonumber \\
\end{gather}
where, $e = u - u_{ref}$. Here, $u_{ref}$ denotes the reference solution, either exact or obtained from standard numerical simulations, while $u$ represents the PINN solution.

For comparison of the trained weight matrices of some layer, say layer $l$, over two runs, say $1$ and $2$, we define:
\begin{gather}
\mathrm{fro}(W^1,W^2) = 
\frac{\left[ \sum_{i,j} \left( W^1_{ij} - W^2_{ij} \right)^2 \right]^{1/2}}
{\left[ \sum_{i,j} \left( W^1_{ij} \right)^2 \right]^{1/2}}, \nonumber \\
\mathrm{cos}(W^1,W^2) = 
\frac{\sum_{i,j} W^1_{ij} W^2_{ij}}
{\left[\sum_{i,j} \left(W^1_{ij} \right)^2\right]^{1/2}
\left[\sum_{i,j} \left(W^2_{ij}\right)^2\right]^{1/2}}, \nonumber \\
\mathrm{corr}(W^1,W^2) =
\frac{
\sum_{i,j}
\left(W^1_{ij}-\overline{W^1}\right)
\left(W^2_{ij}-\overline{W^2}\right)
}
{
\left[
\sum_{i,j}\left(W^1_{ij}-\overline{W^1}\right)^2
\right]^{1/2}
\left[
\sum_{i,j}\left(W^2_{ij}-\overline{W^2}\right)^2
\right]^{1/2}
}
\end{gather}
The relative Frobenius norm measures the normalized difference between weight matrices, and the cosine similarity quantifies their alignment across different training runs. The Pearson correlation calculates the linear dependence of the flattened weight vectors. $\overline{W^1}$ and $\overline{W^2}$ denote the mean values of the entries of the corresponding weight matrices. We now present examples illustrating how randomness can lead to solutions in a manner that is not immediately apparent or easily interpretable, arising from a highly degenerate solution space.

%------------------------------------------------------------------------------------------------%
\begin{figure}[htbp]
    \centering
    \includegraphics[width=0.85\linewidth]{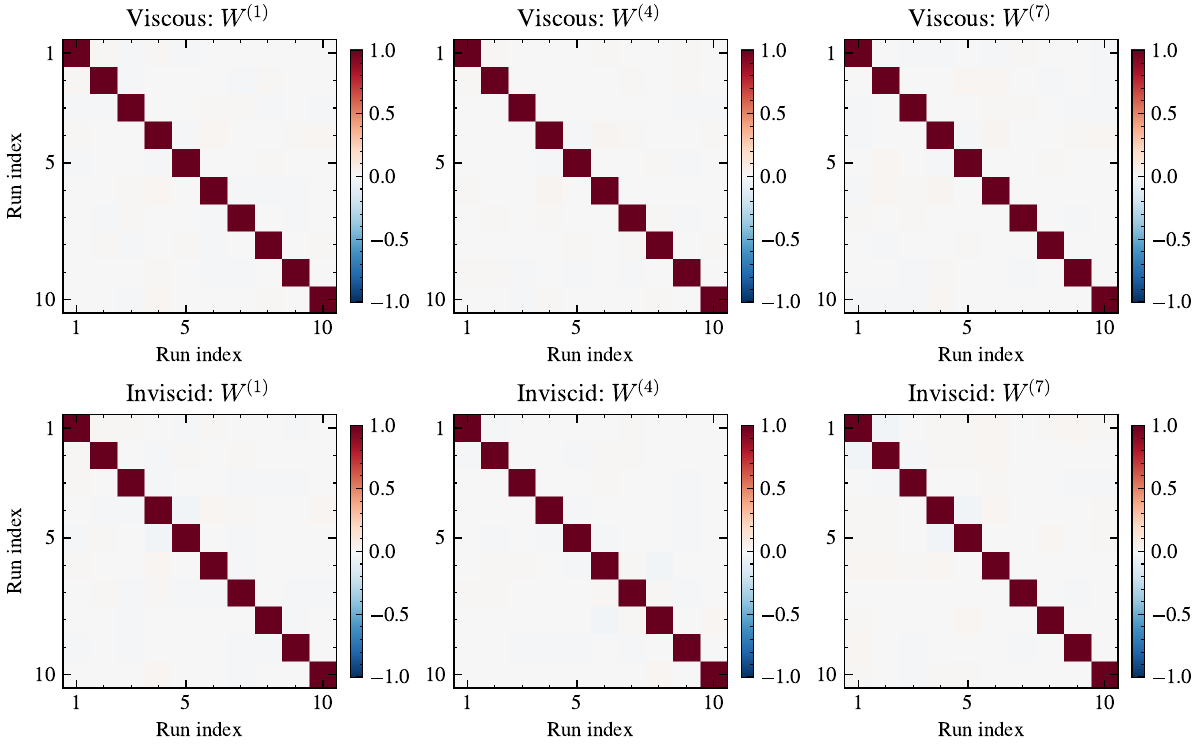}
    \caption{Inter-run Pearson correlation matrices of the trained weights for selected layers across independent PINN realizations for the viscous and inviscid Burgers' equations.
    }
    \label{fig:correlation}
\end{figure}
%------------------------------------------------------------------------------------------------%

%---------------------------------------------------------------------------------%
\subsubsection{Viscous Burgers equation}\label{sec:vis_burg}
%---------------------------------------------------------------------------------%
We first look at the viscous Burgers equation, widely used as a model for nonlinear 
transport and diffusion in reacting and turbulent flows~\cite{frisch1995turbulence, law2006combustion, crank1975mathematics, briscolini1994}. The 1D viscous Burgers' equation for a field $u = u(q) = u(\mathcal{X},\mathcal{T})$ is given as follows:
\begin{equation}
\label{BUR_VIS}
\partial_\mathcal{T} u + u \partial_\mathcal{X} u = \nu\partial^2_\mathcal{XX}u.
\end{equation}

The viscosity $\nu$ is kept as $0.01/\pi$. The initial and boundary conditions for the study are respectively kept as:
\begin{equation}
\label{eq:IC_VIS}
u(\mathcal{X}, 0) = sin\left(2\pi\mathcal{X}\right),
\end{equation}
and,
\begin{equation}
u(0, \mathcal{T}) =  u(1, \mathcal{T}) = 0.
\label{eq:BC_VIS}
\end{equation}

Specifically, Eq.~\ref{eq:chain_update} and Eq.~\ref{eq:propagator} are used for the discrete time marching as shown here. The propagator for PINN $(\omega=1)$ becomes,
\begin{gather}
    T_f[W,b\,;1] = f[W,b].
\end{gather}
This leads to the following non-linear time marching:
\begin{gather}
    \implies z(t+1) = f[W^{(t+1)},b^{(t+1)}]z(t)  = tanh\left(W^{(t+1)}z(t)+b^{(t+1)}\right),
\end{gather} 
where $W$ and $b$ are the trained weights and biases.

The network used here to solve the (both viscous and the inviscid) Burgers' equation, takes a 2 dimensional/feature input $z^{(0)}(q) = \left[\mathcal{X},\mathcal{T}\right]$, passes it 
through $L=8$ hidden layers of $N=100$ neurons each (i.e., a 100 dimensional latent representation), and 
finally maps it to a scalar output $u(\mathcal{X},\mathcal{T})$. 
So effectively,
\begin{equation}
z^{(0)}[\mathcal{X}, \mathcal{T}] \rightarrow z^{(1)}[100] \rightarrow z^{(2)}[100] \rightarrow .\,.\,. \rightarrow z^{(8)}[100] \rightarrow u(\mathcal{X},\mathcal{T}).
\end{equation}

Hence, within the hidden layers, the evolution of $z(q,t)$ can be interpreted as a discrete dynamical system, where both $q$ and $t$ are discrete variables. The use of an equal number of neurons in each layer corresponds to a grid in the discrete space-like variable $q$, whose resolution remains fixed as the system evolves in the depth-representative time $t$.

The PINN-predicted solutions reported in Fig.~\ref{fig:pinn_vs_dyn} show a good agreement with the standard reference numerical solution, also shown in the norms of Tab.~\ref{tab:combined_errors}. All the runs reach essentially the same Loss function minima. This acts as a case where the trajectory points of the signal $z(q, t=L+1)$ for the discrete systems from the runs coincide with each other, and also with a numerically acceptable solution to the problem.

The randomness in weight matrices can be seen from the ensemble representations in Fig.~\ref{fig:weights_burgers_sine}), where the mean and standard deviations show no lower order structures and are rather dense matrices. The same is observed for the individual weight matrices for each run as well.
Furthermore, the independence in each weight matrix is shown quantitatively in Tab.~\ref{tab:weight_comparison} using similarity metrics for a pair of runs as an instance, where a high Frobenius norm and low cosine similarity of $\mathcal{O}~(10^{-2})$ acts as the evidence for the same. A similar trend has been observed for any arbitrary pair of runs. Even more evidence of the lack of any correlations in the trained weights can be found in Fig.~\ref{fig:correlation}, where the correlation matrices of the network layers are identity matrices effectively.

\begin{figure}[t]
    \centering
    \includegraphics[width=\linewidth]{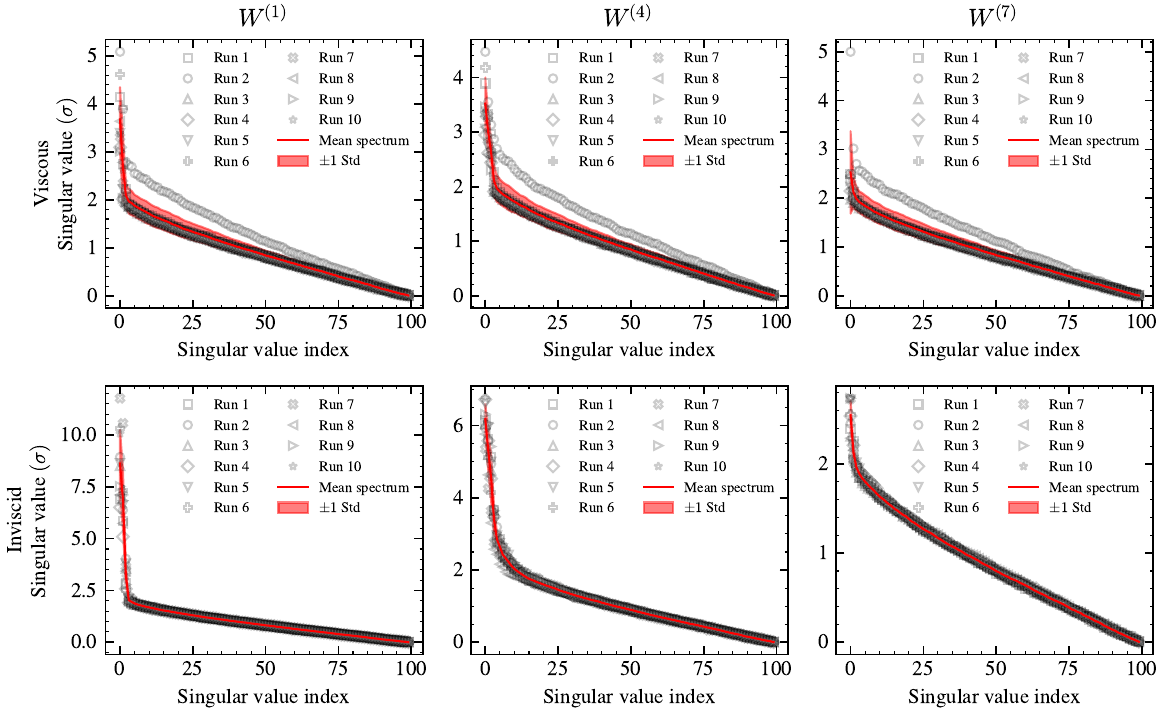}
    \caption{The singular value $(\sigma)$ representation for layers 1,4 and 7 of the PINN solution of the (top) viscous and (bottom) inviscid Burgers' equation for 10 separate, independent runs along with the averages and standard deviations. The spectral representations highlight relative structural similarities in the trained weight matrices across the runs.}
    \label{fig:spectral} 
\end{figure}

However, the spectral representations of trained weight matrices in Fig.~\ref{fig:spectral} show a preferred set of directions in each layer. The singular values $\sigma$ of the trained weight matrices for all the runs are calculated using SVD (Singular Value Decomposition). The variation from the averages, represented using standard deviation regions, remains quite low. In particular, Run 2 seems to have the highest deviation from the average yet settling into the same Loss minima (see Tab.~\ref{tab:combined_errors}).

%------------------------------------------------------------------------------------------------%
\begin{table}[t]
\centering
\small
\begin{tabular}{llccc}
\toprule
\multicolumn{2}{c}{\textbf{Case}} & $L_1$ Error & $L_2$ Error & $L_\infty$ Error \\
\midrule

\multirow{10}{*}{Viscous Burgers'} 
& PINN Run 1  & $1.90\times10^{-3}$ & $4.00\times10^{-3}$ & $2.59\times10^{-2}$ \\
& PINN Run 2  & $1.91\times10^{-3}$ & $4.06\times10^{-3}$ & $2.67\times10^{-2}$ \\
& PINN Run 3  & $1.91\times10^{-3}$ & $4.04\times10^{-3}$ & $2.70\times10^{-2}$ \\
& PINN Run 4  & $1.90\times10^{-3}$ & $4.73\times10^{-3}$ & $4.31\times10^{-2}$ \\
& PINN Run 5  & $1.92\times10^{-3}$ & $4.03\times10^{-3}$ & $2.64\times10^{-2}$ \\
& PINN Run 6  & $1.89\times10^{-3}$ & $4.36\times10^{-3}$ & $3.79\times10^{-2}$ \\
& PINN Run 7  & $1.91\times10^{-3}$ & $4.09\times10^{-3}$ & $3.10\times10^{-2}$ \\
& PINN Run 8  & $1.91\times10^{-3}$ & $4.03\times10^{-3}$ & $2.70\times10^{-2}$ \\
& PINN Run 9  & $1.91\times10^{-3}$ & $4.03\times10^{-3}$ & $2.58\times10^{-2}$ \\
& PINN Run 10 & $1.91\times10^{-3}$ & $4.05\times10^{-3}$ & $2.94\times10^{-2}$ \\

\midrule

\multirow{10}{*}{Inviscid Burgers'} 
& PINN Run 1  & $1.39\times10^{-1}$ & $1.97\times10^{-1}$ & $5.86\times10^{-1}$ \\
& PINN Run 2  & $1.38\times10^{-1}$ & $1.93\times10^{-1}$ & $5.22\times10^{-1}$ \\
& PINN Run 3  & $1.41\times10^{-1}$ & $1.97\times10^{-1}$ & $5.69\times10^{-1}$ \\
& PINN Run 4  & $1.38\times10^{-1}$ & $1.98\times10^{-1}$ & $6.30\times10^{-1}$ \\
& PINN Run 5  & $1.42\times10^{-1}$ & $2.00\times10^{-1}$ & $5.93\times10^{-1}$ \\
& PINN Run 6  & $1.35\times10^{-1}$ & $1.91\times10^{-1}$ & $5.23\times10^{-1}$ \\
& PINN Run 7  & $1.40\times10^{-1}$ & $2.00\times10^{-1}$ & $6.09\times10^{-1}$ \\
& PINN Run 8  & $1.42\times10^{-1}$ & $1.97\times10^{-1}$ & $5.12\times10^{-1}$ \\
& PINN Run 9  & $1.35\times10^{-1}$ & $1.96\times10^{-1}$ & $6.14\times10^{-1}$ \\
& PINN Run 10 & $1.41\times10^{-1}$ & $1.97\times10^{-1}$ & $5.85\times10^{-1}$ \\

\midrule

\multirow{10}{*}{Eikonal} 
& PINN Run 1  & $2.12\times10^{-5}$ & $3.85\times10^{-5}$ & $7.04\times10^{-5}$ \\
& PINN Run 2  & $1.03\times10^{-4}$ & $1.14\times10^{-4}$ & $1.89\times10^{-4}$ \\
& PINN Run 3  & $4.58\times10^{-5}$ & $6.40\times10^{-5}$ & $1.82\times10^{-4}$ \\
& PINN Run 4  & $1.26\times10^{-4}$ & $3.34\times10^{-4}$ & $2.61\times10^{-3}$ \\
& PINN Run 5  & $2.74\times10^{-4}$ & $7.07\times10^{-4}$ & $4.80\times10^{-3}$ \\
& PINN Run 6  & $3.67\times10^{-5}$ & $4.87\times10^{-5}$ & $1.18\times10^{-4}$ \\
& PINN Run 7  & $4.93\times10^{-5}$ & $6.66\times10^{-5}$ & $1.14\times10^{-4}$ \\
& PINN Run 8  & $6.05\times10^{-5}$ & $7.47\times10^{-5}$ & $1.25\times10^{-4}$ \\
& PINN Run 9  & $2.75\times10^{-1}$ & $3.18\times10^{-1}$ & $4.84\times10^{-1}$ \\
& PINN Run 10 & $7.86\times10^{-5}$ & $1.15\times10^{-4}$ & $3.12\times10^{-4}$ \\
\bottomrule
\end{tabular}

\caption{
Error comparison of PINN solutions obtained from independent runs across different equations against reference solutions at $t=0.3\,s$ over a uniform spatial grid of 100 test points.
}

\label{tab:combined_errors}
\end{table}
%------------------------------------------------------------------------------------------------%

\begin{table}[t]
\centering
\small
\begin{tabular}{ccccc}
\toprule
& \multicolumn{2}{c}{\textbf{Viscous}} & \multicolumn{2}{c}{\textbf{Inviscid}} \\
\textbf{Layer} 
& $\mathrm{fro}(W^1,W^2)$ & $\mathrm{cos}(W^1,W^2)$ 
& $\mathrm{fro}(W^1,W^2)$ & $\mathrm{cos}(W^1,W^2)$ \\
\midrule
0  & $6.696 $ & $0.0813$  & $1.473 $ & $0.0124$ \\
1  & $1.695 $ & $0.0153$  & $1.488 $ & $0.0139$ \\
2  & $1.628 $ & $-0.0030$ & $1.439 $ & $-0.0190$ \\
3  & $1.622 $ & $-0.0023$ & $1.415 $ & $0.0091$  \\
4  & $1.631 $ & $0.0110$  & $1.427 $ & $-0.0130$\\
5  & $1.674 $ & $-0.0106$ & $1.404 $ & $-0.0000$ \\
6  & $1.694 $ & $-0.0077$ & $1.391 $ & $0.0228$  \\
7  & $1.733 $ & $0.0050$  & $1.431 $ & $-0.0235$\\
8  & $1.280 $ & $0.1465$  & $1.254 $ & $0.1837$ \\
\bottomrule
\end{tabular}
\caption{Layerwise comparison of weight matrices across independent PINN training runs for inviscid and viscous cases, using relative Frobenius norm and cosine similarity.}
\label{tab:weight_comparison}
\end{table}

From the lens of interpretability, the distinct and largely uncorrelated weight matrices obtained across independent PINN runs indicate that the learned dynamics do not recover the structured operators typically associated with grid-based discretizations of PDEs, where information is concentrated in a few entries (e.g., tridiagonal matrices). On the contrary, it has been shown~\cite{tucny2026randomness} that, for this class of problems, the trained weights exhibit an approximately random, near-Gaussian distribution, both in their entries and in their spectral properties. The representation appears to be distributed across the parameter space. From the current underconstrained Loss function perspective, this behavior reflects the non-uniqueness of the weights in the optimization process, whereby both structured and unstructured representations can yield comparable solutions, with the resulting distributed representation offering robustness at the expense of interpretability. This is consistent with the dynamical systems interpretation as well, wherein the mapping between parameters ${W,b}$ and the resulting solution $z(q,t=L+1) = u(\mathcal{X},\mathcal{T})$ is highly degenerate, allowing different parameter realizations to yield similar representations in function space, even if their underlying dynamics differ.

%--------------------------------------------------------------------------%
%--------------------------------------------------------------------------%

 %---------------------------------------------------------------------------------%
 \subsubsection{Inviscid Burgers equation}\label{sec:invis_burg}
 %---------------------------------------------------------------------------------%
The fact that random matrices are astronomically more numerous than tridiagonal ones
does not per se guarantee that a PINN will always find a suitable solution. 
To further inspect this point, we next consider the inviscid form of the Burgers equation, a prototypical model for shock formation 
and nonlinear wave propagation~\cite{whitham1974linear, leveque2002finite}:
\begin{equation}
\label{BUR}
\partial_\mathcal{T} u + u \partial_\mathcal{X} u =0.
\end{equation}
We take a Riemann type initial condition, which is intentionally 
discontinuous, as it is known that traditional PINNs struggle to capture shocks without additional safeguards, consistent with known gradient pathologies and optimization difficulties of PINNs~\cite{wang2021understanding, krishnapriyan2021characterizing}.
The initial and boundary conditions are given by:
\begin{equation}
u(\mathcal{X}, 0) = 
\begin{cases} 
    1, & \mathcal{X} < 0.5, \\
    0, & \mathcal{X} \ge 0.5 \qquad ,
\end{cases}
\label{eq:IC}
\end{equation}
and,
\begin{equation}
u(0, \mathcal{T}) = 1, \qquad u(1, \mathcal{T}) = 0.
\label{eq:BC}
\end{equation}

%------------------------------------------------------------------------------------------------%
\begin{figure}[htbp]
    \centering
    \includegraphics[width=0.9\linewidth]{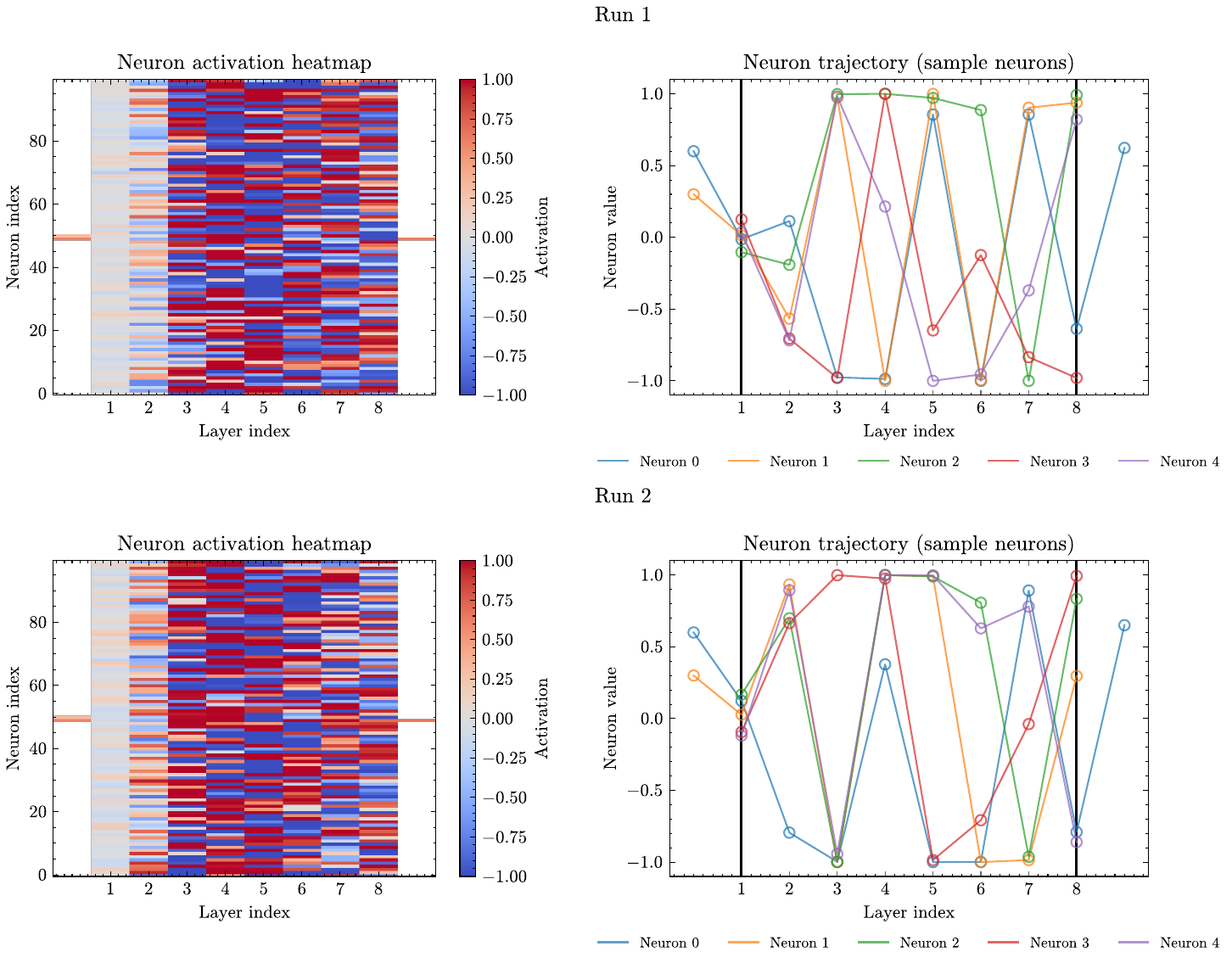}
    \caption{The figure shows the evolution of a 2D input signal as it propagates through the PINN architecture trained on the inviscid Burgers’ equation, across two independently initialized runs. The left panel displays a heatmap of neuron activations across layers, while the right panel illustrates the trajectories of five representative neurons, 
    $z(q,t)$, within the 100-dimensional latent space for the input 
    $z^{(0)} = x= [0.6,0.3]$. The input and output layers are indicated schematically at the center for visualization purposes.}
    \label{fig:neurontrajectory_burgers_riemann}
\end{figure}
%------------------------------------------------------------------------------------------------%

%------------------------------------------------------------------------------------------------%
\begin{figure}[htbp]
    \centering
    \includegraphics[width=\linewidth]{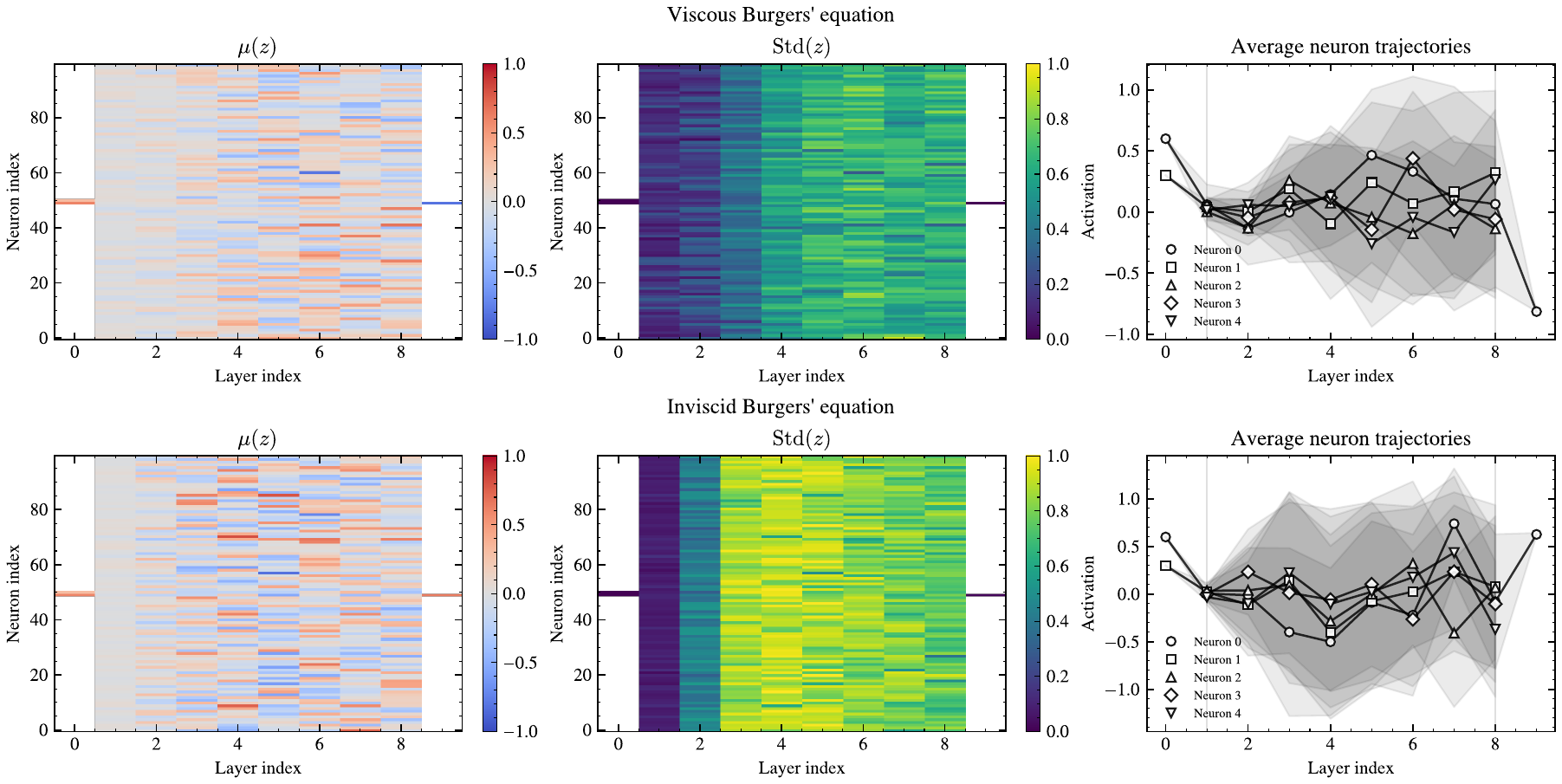}
    \caption{
    Ensemble activation statistics across network layers for 10 independently trained PINN realizations of the viscous and inviscid Burgers' equations. The mean $\mu(z)$ and standard deviation $Std(z)$ activation heatmaps are shown together with representative average neuron trajectories across layers.
    }
    \label{fig:activation_average}
\end{figure}
%------------------------------------------------------------------------------------------------%

The present network, however, fails to learn the problem in all the runs, as seen in Fig.~\ref{fig:pinn_vs_dyn}. In each case, the discrete trajectories arising out of the corresponding neural PDEs, pass through to the same point in the Loss function landscape at $t=L+1$, albeit an incorrect one. While the predicted solutions from both runs are nearly identical, they are not numerically acceptable, suggesting a more rugged error landscape in high dimensions. Ensemble representations of the trained weights in Fig.~\ref{fig:weights_burgers_sine} show a similar lack of structures analogous to the viscous case.

The degeneracy in the learning procedure is explicitly demonstrated through the evolution of $z(q,t)$ across the deep layers for an 2-feature input signal at $\left[\mathcal{X}=0.6, \mathcal{T}=0.3\right]$, interpreted via the discrete dynamical system analogy (see the right panel of Fig.~\ref{fig:neurontrajectory_burgers_riemann}) for two separate runs. For clarity, only 5 representative neuron trajectories are shown out of 100 per deep layer. Across independent runs, the activation of a given neuron follows markedly different paths, reflecting differences in the learned weights. The extent of difference in the two dynamical systems as a result of different weights needs further investigation. These differences originate from the first hidden layer itself, where the input is projected into a higher-dimensional latent space, leading to distinct initial conditions for subsequent evolution. However, starting from the same input, the trajectories of different runs collapse to essentially the same, incorrect output.

Average activation heatmaps for the viscous and inviscid Burgers' equations are shown in Fig.~\ref{fig:activation_average} over 10 independent runs. The hidden layers exhibit the strongest activation variability, as indicated by the elevated standard deviation regions, suggesting that the latent representations are most expressive in these layers. For the inviscid Burgers' equation, the activation statistics exhibit substantially sharper layer-to-layer variations, particularly in the standard deviation activation maps. While strong neural activity is still observed within the deeper latent layers, the early hidden layers display comparatively weaker and less spatially distributed activation patterns. This behavior suggests a less coherent propagation of informative features through the network depth, with abrupt changes in latent feature amplification and suppression across layers. Such irregular internal representation dynamics may be associated with the optimization difficulties commonly encountered in shock-dominated hyperbolic problems. An analysis to determine the cause of this non-learnability by the network is not performed in the present context. In contrast, the viscous case exhibits comparatively smoother activation statistics and more spatially distributed activation patterns in the early hidden layers (see $\operatorname{Std}(z)$ near layer index 1), indicating a more stable and expressive latent representation evolution through the network depth.

The neuron trajectory plots in Fig.~\ref{fig:activation_average} additionally demonstrate substantial variability across neurons and layers, suggesting significant diversity in the internal representations learned by independently trained network realizations. From the dynamical systems perspective of deep neural networks, these trajectories may be interpreted as distinct latent evolution pathways through the network depth, indicating the existence of multiple admissible internal solution trajectories capable of producing comparable macroscopic predictions.

Finally, the similarity metrics in Tab.~\ref{tab:weight_comparison} quantify the non-uniqueness of the learned weights, which is consistent with the viscous case. However a closer spectral similarity in the trained weights of independent runs can be observed for the inviscid case in Fig.~\ref{fig:spectral}, when compared to the viscous case even though the true solution is never reached satisfactorily by any of the realizations.

%--------------------------------------------------------------------------%
\begin{figure}[t]
\centering
\includegraphics[width=0.4\linewidth]{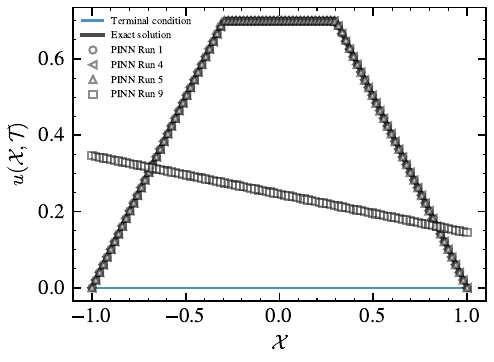}
\caption{The figure compares the solutions obtained by two PINN runs for the 1D 
Eikonal equation with the exact solution at time $t = 0.3$. Run 9 fails to learn the exact solution satisfactorily.}
\label{fig:solution_eikonal}
\end{figure}
%--------------------------------------------------------------------------%

%---------------------------------------------------------------------------------%
\subsection{Eikonal Equation}
%---------------------------------------------------------------------------------%
We then consider the one-dimensional backward Eikonal equation, that appears in several areas of chemical physics, including semi-classical descriptions of molecular scattering and reaction dynamics~\cite{sethian1999levelset, miller1970semiclassical}, propagation of reaction fronts in heterogeneous media~\cite{wilder1993modification}, and thermodynamic formulations based on gradient flows and information geometry~\cite{wada2021eikonal}. It is given by:
\begin{equation}
    -\partial_\mathcal{T} u(\mathcal{X},\mathcal{T}) + | \, \partial_\mathcal{X} u(\mathcal{X},\mathcal{T}) \,| = 1, 
    \qquad \left(\mathcal{X},\mathcal{T}\right) \in [-1,1] \times [0,1),
\end{equation}
and for the present study, supplemented with a terminal condition at final time $\mathcal{T}_f=1$,
\begin{equation}
    u(\mathcal{X},\mathcal{T}_f) = 0, \qquad \mathcal{X} \in [-1,1],
\end{equation}
and homogeneous Dirichlet boundary conditions along the spatial domain,
\begin{equation}
    u\left(-1,\mathcal{T}\right) = u\left(1,\mathcal{T}\right) = 0, 
    \qquad \mathcal{T} \in \left[0,\mathcal{T}_f\right).
\end{equation}

%--------------------------------------------------------------------------%
\begin{figure}[t]
    \centering
    \includegraphics[width=\linewidth]{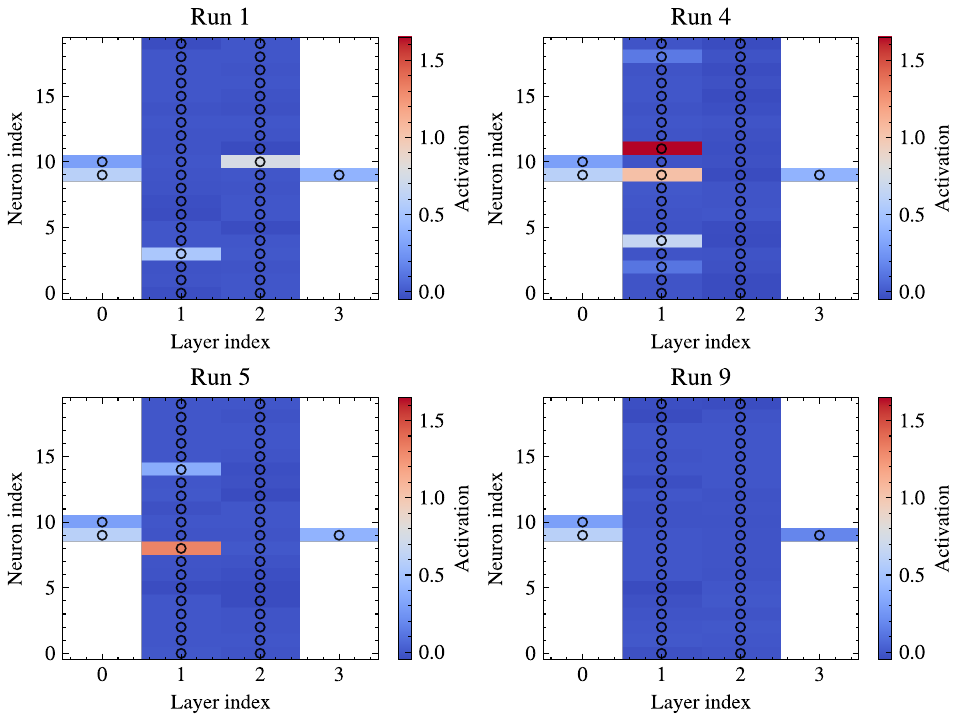}
    \caption{The figure shows shows the heatmap of the activation values of each neuron in the network obtained after learning the 1D Eikonal equation, for an input signal $z^{(0)} = x= [0.6,0.3]$ for 4 independent training realizations. Input and output layers are representatively shown in the center instead of their corresponding exact neuron indices. Run 9 with relatively non-expressive activations fails to learn the true solution.}
    \label{fig:activation_eikonal}
\end{figure}
%--------------------------------------------------------------------------%

This problem corresponds to a time–reversed Hamilton–Jacobi equation 
in which the solution $u(\mathcal{X},\mathcal{T})$ represents the accumulated travel time from the 
point $(\mathcal{X},\mathcal{T})$ to the final time $\mathcal{T}_f$ under a unit-speed metric. 
Note that the problem is still inviscid, but with continuous initial and boundary conditions. Therefore, this problem acts as an intermediate level of learning problem, placed between the viscous and inviscid versions of the Burgers' equation.

A small network is used for this problem, with just 2 deep layers, each with 20 neurons each. The activation function used here is Leaky ReLU (with a negative slope of 0.01). The learning rates are kept the same as here~\cite{blechschmidt2021three}. A total of $10,000$ random collocation points are used as training data, trained over $10,000$ epochs.

Similar to the case shown in Sec.~\ref{sec:vis_burg}, 10 separate training runs are performed on the network. Here, the 2 dimensional input evolves in a $20$ dimensional latent space inside the DNN to eventually collapse into a scalar output. Despite the lack of diffusion, most training runs for this network are able to learn the problem fairly well, except for Run 9 where the learning is significantly poor (see Fig.~\ref{fig:solution_eikonal}). 
Despite being about $20\%$ the size of the network 
used for the Burgers' equation both in depth and number of neurons per layer, still follows different paths to optimization with each run, visible in the neural activations shown in Fig.~\ref{fig:activation_eikonal}, albeit not all the trajectories lead to a satisfactory solution. The PINN solution of the Eikonal equation is therefore a demonstration where the resulting discrete dynamical system is more susceptible to relax to a fixed point without ever reaching a satisfactory Loss function minima with respect to the problem PDE itself. This problem also provides a concrete numerical demonstration of the existence of multiple largely independent relaxation trajectories followed by the underlying discrete dynamical system, at least one of which never crosses an optimum Loss function mimima. The learning inaccuracy of Run 9 is quantified in Tab.~\ref{tab:combined_errors}.

It can be observed in Fig.~\ref{fig:activation_eikonal} that a few specific neurons behave in significantly different ways across runs in evaluating the solution. This seemingly special subset of neurons may be speculatively seen as an analogy to the so-called feature neurons~\cite{DecodingFeatureNeurons2025}, which play a key role in understanding how Large Language Models (LLMs) learn different concepts. However, it is also observed that these feature-like neurons are not unique and vary from run to run. This indicates that no neuron has an intrinsic functional role tied to the problem. Instead, different runs assign the dominant contributions to different neurons, highlighting the existence of multiple distinct pathways to learning the same solution in a highly over-parameterized space. Finally, Run 9, corresponding to the failed learning case, exhibits the weakest neural activations among all realizations in Fig.~\ref{fig:activation_eikonal}, lacking any aforementioned ``feature neurons" of significance.

%---------------------------------------------------------------------------------%
\section{Link to neural PDEs and neural integral equations}
%---------------------------------------------------------------------------------%
The formulation in Sec.~\ref{sec:DDS} establishes a connection between neural networks and differential operators. In this framework, the convolution with the kernel $W$ governs the evolution of the state in the associated discrete dynamical system. This suggests that, under suitable assumptions where the kernel exhibits structured behavior, it may be possible to relate $W$ to an underlying neural PDE. To make this link explicit, we rewrite the
convolution in terms of a shifted coordinate $r = q' - q$:
\begin{equation}
\bar{z}(q,t) = \int W(q,q+r,t)\, z(q+r,t)\, dr.
\end{equation}
Assuming sufficient smoothness, a Taylor expansion around $r=0$ yields
\begin{equation}
\bar{z}(q,t) = \sum_{k=0}^{k_{max}} W_k(q,t)\, \partial_q^k z(q,t),
\end{equation}
where the coefficients
\begin{equation}
W_k(q,t) = \int W(q,r,t)\, r^k\, dr
\end{equation}
represent the spatial moments of the kernel.

Substituting into Eq.~\ref{REL}, the dynamics can be expressed as a
neural PDE:
\begin{equation}
\label{NPDE}
z_t = -\gamma \big(z - f \left(W_0 z + W_1 z_q + W_2 z_{qq} + \dots - b \right) \big).
\end{equation}

In this representation, even-order terms $W_{2k}$ correspond to generalized
diffusion processes, while odd-order terms $W_{2k+1}$ describe advective or
propagative effects. Hence, the structure of the kernel $W(q,q',t)$ directly
determines the type and order of the underlying differential operator.

In the linear case $f(z)=z$, Eq.~\ref{NPDE} reduces to a standard PDE.
For example, retaining only first and second-order terms leads to an
advection-diffusion equation, whose discrete counterpart corresponds
to a tridiagonal weight matrix:
\begin{equation}
\label{eq:weights_adv_diff}
W_{i,j} = \left(D+\frac{U}{2}\right)\delta_{i-1,j}
+ \left(1-2D\right)\delta_{i,j}
+ \left(D-\frac{U}{2}\right)\delta_{i+1,j}.
\end{equation}

This establishes a direct correspondence between the structure of the
weight matrix and the form of the governing differential operator.
In particular, local PDEs are associated with sparse, banded matrices,
while non-local interactions lead to dense or slowly decaying kernels. 
From this perspective, one may expect that training procedures aimed at
learning physical systems, such as Physics-Informed Neural Networks (PINNs), 
would yield structured weight matrices reflecting the underlying operators. 

However, as shown, this is generally not observed: trained
PINNs tend to produce weight matrices with no explicit
low-order structure, often resembling random matrices. Furthermore, each independent training instance of a PINN leads to a completely different arrangement of parameters, only similar from a spectral viewpoint and no guarantee of a valid solution. This suggests that the underlying operator
representation is distributed across the network in a non-trivial way,
rather than being directly encoded in a sparse, structured form, acting as a motivation for the
need for alternative formulations capable of embedding operator-level
information more explicitly within the learning process. Conversely, the statistical properties of $W$ may provide insight into the effective neural PDEs governing the internal dynamics of the network, which could prove useful for improving DNN optimization strategies.

%---------------------------------------------------------------------------------%
\section{Outlook}
%---------------------------------------------------------------------------------%
In summary, we have introduced deep neural networks (DNNs) as discretized realizations of neural integral equations in relaxation form, with particular emphasis on the non-uniqueness (degeneracy) of solutions in Physics-Informed Neural Networks (PINNs). This viewpoint provides evidence that PINNs and classical finite-difference (FD) schemes represent two distinct computational pathways to approximate the same underlying dynamics of a physical system.

From a dynamical systems perspective, the learning process in DNNs can be interpreted as a finite-time evolution governed by discrete neural PDEs, which may be viewed as an attempt to approach underlying attractors. However, in finite-layer PINNs, the training procedure enforces agreement with the target solution only at the final layer, without explicitly constraining convergence to the corresponding attractor of the underlying dynamics. As a result, the internal dynamical structure associated with these attractors is only partially captured, in contrast to formulations that directly target equilibrium or continuous-depth limits.

We also highlight a fundamental degeneracy in the inverse mapping between network parameters and learned representations, with multiple parameter configurations leading to comparable solutions. In contrast to structured discretizations, where operators are explicitly defined, PINNs rely on distributed representations across high-dimensional parameter spaces, which may offer increased flexibility but at the cost of interpretability and computational expense.

The analogy between DNNs, neural integral equations, and their discrete counterparts suggests that learning dynamics in feed-forward networks can be studied within the broader framework of computational physics. This perspective opens the door to systematic investigations of stability, convergence, and representation in neural architectures, and may extend to more general architectures with non-local or non-Markovian interactions.

An important open question concerns the effective ``spacing'' of features as they evolve across network depth. While the present formulation assumes a uniform discretization, there is no fundamental reason for this to hold in general, and extensions to non-uniform or adaptive representations may provide a more accurate description of the underlying dynamics.

Finally, in high-dimensional settings where classical grid-based methods become impractical, stochastic approaches play a central role in computational statistical physics. Extending the present dynamical systems framework to stochastic dynamics remains an open direction, with potential connections to emerging approaches that combine machine learning with path-sampling techniques for complex systems~\cite{Jung2023Dellago}.

%---------------------------------------------------------------------------------%
\section{Acknowledgements}
%---------------------------------------------------------------------------------%

S. S. is grateful to SISSA for financial support under the
``Collaborations of Excellence" initiative as well as 
to the Simons Foundation for supporting several enriching visits of his.
He also wishes to acknowledge many enlightening discussions 
over the years with P. V. Coveney, A. Laio and D. Spergel.
S. S. acknowledges funding from the from the Research and Innovation programme ``European Union’s
Horizon Europe EIC pathfinder'' under grant agreement ``No101187428''.

S. A. acknowledges support from the Abdul Kalam Technology Innovation National Fellowship (INAE/SA/4784).

%---------------------------------------------------------------------------------%
\section{Dedication} 
%---------------------------------------------------------------------------------%
This work is dedicated to Chris Dellago, a highly esteemed colleague and a 
lifelong friend of S. S., with the warmest wishes of great continued success 
for many years to come. 

%---------------------------------------------------------------------------------%
\section{Author Declarations}
%---------------------------------------------------------------------------------%
\subsection{Conflict of Interest}
The authors have no conflicts to disclose.

%---------------------------------------------------------------------------------%
\section{Data availability statement}
%---------------------------------------------------------------------------------%
The scripts and datasets required to reproduce the results presented in this work can be found in the following GitHub repository: \url{https://github.com/abhishekganguly808/degeneracy-in-pinns}.
%---------------------------------------------------------------------------------%
%---------------------------------------------------------------------------------%
\bibliographystyle{unsrt}  
\bibliography{bibML_Dellago}

\end{document}